\documentclass[letterpaper, 10 pt, conference]{ieeeconf}
\IEEEoverridecommandlockouts
\usepackage{graphicx}
\usepackage{amsmath,amssymb,amsfonts}
\usepackage{bm}
\usepackage{mathtools}
\usepackage{siunitx}
\usepackage{booktabs}
\usepackage{multirow}
\usepackage{placeins} 
\usepackage{dblfloatfix} 
 
\usepackage{algorithm}
\usepackage{tabularx}
\usepackage{array}
\newcolumntype{C}{>{\centering\arraybackslash}X}
\usepackage{algorithmic}
\usepackage{hyperref}
\usepackage{xcolor}
\usepackage{enumitem}
\usepackage{microtype}
\usepackage{nicefrac}
\usepackage{wrapfig}
\usepackage{caption}
\usepackage{subcaption}
\usepackage{placeins}
\usepackage{float}
\usepackage{dsfont}
\usepackage{pifont}
\usepackage{array}
\usepackage{url}
\usepackage{cite}
\usepackage{bbm}
\usepackage{booktabs} 

\usepackage{amsthm}
\usepackage{cuted} 
\usepackage{afterpage}

\newtheorem{theorem}{Theorem}[section]

\title{Dense-Jump Flow Matching with Non-Uniform Time Scheduling for Robotic Policies: Mitigating Multi-Step Inference Degradation}

\author{
  Zidong Chen$^{1}$, Zihao Guo$^{2}$, Peng Wang$^{3}$$^{*}$, ThankGod Itua Egbe$^{2}$, Yan Lyu$^{4}$, Chenghao Qian$^{5}$
  \thanks{$^{1}$Z. Chen is with Dept. Computing, Imperial College London}
  \thanks{$^{2}$Z. Guo \& T. Egbe are with Dept. Computing, Manchester Metropolitan University}
  \thanks{$^{3}$P. Wang is with CVSSP, University of Surrey}
  \thanks{$^{4}$Y. Lyu is with School of Computer Science and Engineering, Southeast University}
  \thanks{$^{5}$C. Qian is with Institute for Transport
Studies, the University of Leeds}
  \thanks{$^{*}$Corresponding author}}

\begin{document}
\maketitle

\thispagestyle{empty}
\pagestyle{empty}
\begin{abstract}

Flow matching has emerged as a competitive framework for learning high-quality generative policies in robotics; however, we find that generalisation arises and saturates early along the flow trajectory, in accordance with recent findings in the literature. We further observe that increasing the number of Euler integration steps during inference counter-intuitively and universally degrades policy performance. We attribute this to (i) additional, uniformly spaced integration steps oversample the late-time region, thereby constraining actions towards the training trajectories and reducing generalisation; and (ii) the learned velocity field becoming non-Lipschitz as integration time approaches 1, causing instability. To address these issues, we propose a novel policy that utilises non-uniform time scheduling (e.g., U-shaped) during training, which emphasises both early and late temporal stages to regularise policy training, and a dense-jump integration schedule at inference, which uses a single-step integration to replace the multi-step integration beyond a jump point, to avoid unstable areas around 1. Essentially, our policy is an efficient one-step learner that still pushes forward performance through multi-step integration, yielding up to 23.7\% performance gains over state-of-the-art baselines across diverse robotic tasks. Code is open-sourced at \url{https://github.com/DenseJumpFM/DenseJump_FlowMatching}


\end{abstract}

\section{Introduction}



\afterpage{
    \begin{figure*}[!htbp]
        \centering
        \vspace{-0.5em}
        \includegraphics[width=1.0\textwidth]{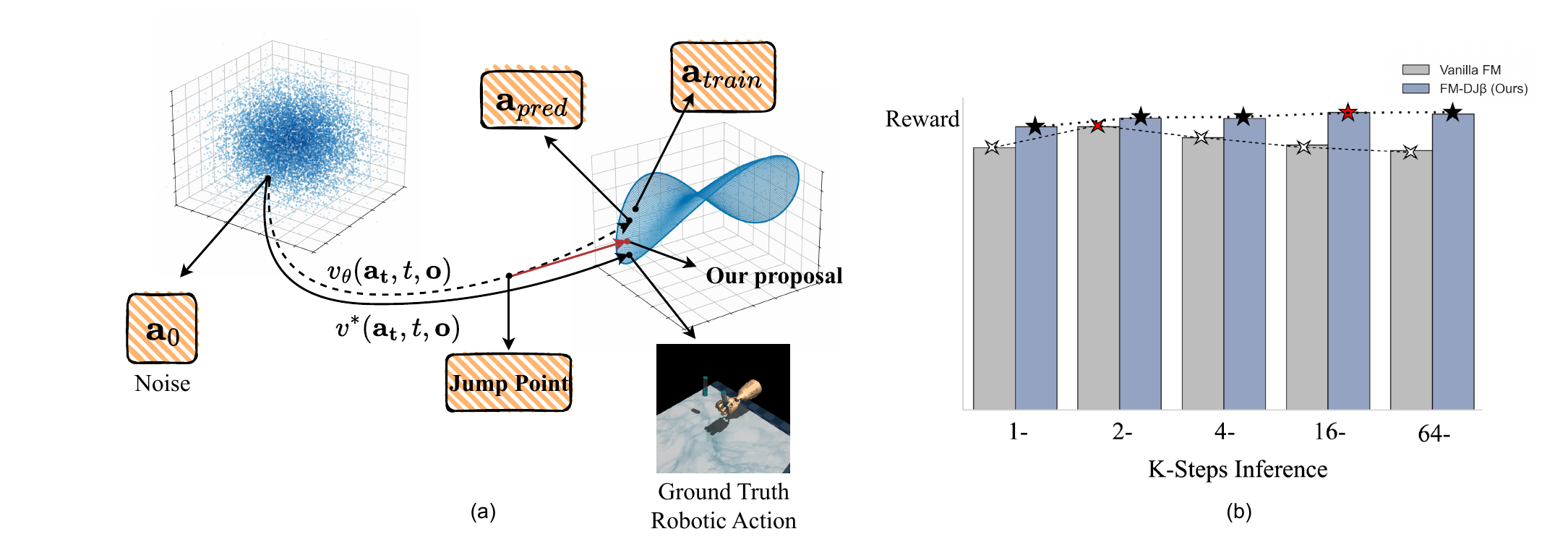}
        \caption{(a) Comparison of vanilla flow matching policy and the proposed dense-jump flow matching (FM-DJ$\beta$) policy. A standard flow matching policy starts from Gaussian noise $\mathbf{a}_0$ and, by learning a velocity field $v_\theta(\mathbf{a}_t, t, \mathbf{o})$, produces an action trajectory conditioned on observation $\mathbf{o}$. The ideal target velocity $v^*(\mathbf{a}_t, t, \mathbf{o})$ would transport $\mathbf{a}_0$ directly to the expert action. In practice, $v_\theta$ drifts toward training actions, undermining generalisation. FM-DJ$\beta$ introduces non-uniform time sampling during training plus a dense-jump integration schedule at inference, stabilising late-time behaviour and improving the predicted action $\mathbf{a}_{\text{pred}}$. (b) Walker2D average reward vs.\ inference steps: vanilla FM degrades as steps increase, while FM-DJ$\beta$ maintains strong performance across step counts.}
        \label{fig:fm_policy}
        \vspace{-0.75em}
    \end{figure*}
}

Generative policies have gained substantial momentum in robot learning due to their success in other areas~\cite{ho2022video,black2024pi0,chi2023rss,zhang2024affordance,wang2024robot}. Compared to traditional formulations that treat policy learning as supervised regression from states to actions, generative approaches learn the full multimodal conditional distribution, capturing the inherent ambiguity of many robotic tasks. Among the approaches, diffusion-based generative methods have been proved to be theoretically bounded and practically applicable. For instance, the diffusion policy~\cite{chi2023rss} learns to denoise actions through a time-indexed Markov process, which has been demonstrated to be successful through tasks like push-T. Diffusion models bring two key benefits: (i) they cover diverse modes of the expert action distribution, and (ii) their training objective, i.e., the denoising/score matching, often yields stable optimisation from large, off-policy datasets. Recent works further show that these models scale effectively to high-dimensional observations and action spaces, yielding strong performance across diverse benchmarks~\cite{ze2024rss, zhao2023rss,chi2023rss,wang2024robot,rouxel2024flow}. However, iterative denoising typically requires extra inference steps, leading to latency and introducing sensitivity to the timestep schedule, which limits its application in real-time robotic tasks.



Flow Matching (FM) offers a faster alternative to diffusion: instead of a stochastic denoising process, FM learns a deterministic Ordinary Differential Equation (ODE) velocity field that transports simple noise to expert-like actions via integration. In principle, FM supports both multi-step and even one-step sampling, narrowing the training–inference gap and reducing latency, making it highly promising for real-time robotics. Recent studies have adapted FM to robotic policy learning, demonstrating that FM achieves comparable performance to diffusion while being significantly faster~\cite{zhang2025flowpolicy,zhang2024robot}.

The appeal of FM-based robotic policies lies in their ability to leverage one- or multi-step integration for fast inference. However, several challenges remain. (i) One-step inference can fail, and resampling from the source distribution rarely improves performance;
(ii) Multi-step inference, while less efficient, can progressively refine trajectories, correcting errors from earlier steps and increasing the chance of recovery in difficult scenarios. Thus, accurate multi-step inference is valuable for robustness, even if it comes at the cost of some efficiency. These observations motivate the need for FM policies that can flexibly trade off between accuracy and efficiency by adjusting the number of inference steps.

Another overlooked aspect is that FM was originally developed for image generation, not robotic control. It is therefore important to examine the differences between these domains to design effective FM-based policies for robotics. Two key distinctions are: (i) Target-space dimensionality: image generators output high-dimensional tensors (e.g., CIFAR-10~\cite{abouelnaga2016cifar}: $32{\times}32{\times}3$; ImageNet~\cite{deng2009imagenet}: $224{\times}224{\times}3$), whereas robotic policies operate in much lower-dimensional action spaces (e.g., 6–8 DOF grippers, 16–30 DOF dexterous hands, 12–50 DOF humanoids); (ii) Data regimes: image generation benefits from abundant online data, while real-world robotic data collection is expensive, requiring physical trials, supervision, safety compliance, and resets. These differences lead to a counterintuitive finding we document in this paper: FM policy performance peaks at a small number of solver steps and then degrades as steps increase, contrary to the expectation that finer integration should always improve results.

In this work, we provide a principled theoretical explanation for the observed degradation in FM policy performance with increasing inference steps. We first show that the learned velocity field loses local Lipschitz continuity as $t \to 1$, consequently, the Picard–Lindelöf theorem no longer applies, solutions become sensitive and non-unique, amplifying errors and harming generalisation. To overcome these challenges, we introduce: (i) a non-uniform time schedule (e.g., U-shaped) during training that emphasises both early and late integration times, and (ii) a Dense-Jump inference strategy that allocates computation to stable regions and performs a single jump to $t=1$ to avoid instability. Our approach delivers robust few-step performance and enables flexible trade-offs between computational cost and accuracy, directly addressing a key limitation of current FM-based robotic policies. Our main contributions are:
\begin{enumerate}
    \item We present the first systematic study across diverse robotic tasks demonstrating that increasing the number of inference steps in FM policies can degrade performance, contrary to common intuition. 
    \item Through rigorous theoretical and empirical analysis, we identify two root causes: (i) the velocity field becomes non-Lipschitz as $t \to 1$, leading to instability; and (ii) the learned velocity field, under standard training regimes, drifts toward training actions rather than the true expert actions.
    \item We propose two complementary solutions: (i) a non-uniform time sampling schedule during training that encourages robust one- or few-step action generation; and (ii) a Dense-Jump ODE solver that enables multi-step inference to trade compute for accuracy without suffering from late-time instability. We validate these methods across tasks of varying difficulty, achieving up to 23.7\% improvement over state-of-the-art baselines.
\end{enumerate}

\section{Related Work}
Flow matching~\cite{liu2022flow,lipman2023flow} has emerged as a powerful generative modelling framework, leveraging optimal transport theory to learn a velocity field that transports samples from a simple source distribution to a complex target distribution. Unlike diffusion models, which denoise data step by step, FM directly learns the velocity field and generates samples by solving an ODE, resulting in greater efficiency and flexibility. FM has demonstrated strong performance across various domains, including image generation~\cite{liu2022flow,bertrand2025closed} and robotics~\cite{zhang2024affordance,hu2024adaflow,zhang2025flowpolicy}.

Recent work has adapted FM to robotic policy learning, showing its effectiveness for learning from expert demonstrations. A key focus has been improving inference efficiency to enable real-time robotic applications. For example, Zhang et al.~\cite{zhang2024affordance} applied FM to a range of robotic tasks, demonstrating faster inference than diffusion-based policies while maintaining comparable generalisation. Building on the connection between the conditional variance of the training loss and ODE discretisation error, AdaFlow~\cite{hu2024adaflow} introduces a variance-adaptive solver that adjusts step size based on the uncertainty of the learned velocity field, improving both efficiency and performance. FlowPolicy~\cite{zhang2025flowpolicy} proposes consistency flow matching, a generalised FM method that learns segmentwise straight-line flows, enabling one-step inference and balancing efficiency with effectiveness. However, these methods primarily focus on applying FM to robotics or improving inference efficiency, while overlooking the potential performance degradation caused by multi-step inference of FM, which can undermine policy generalisation~\cite{bertrand2025closed}.


\begin{table*}[t]
\centering
\label{tab:inference-steps}
\begin{tabularx}{\textwidth}{l*{5}{C}}
\toprule
\textbf{Inference steps} & \textbf{1 Step} & \textbf{2 Steps} & \textbf{4 Steps} & \textbf{16 Steps} & \textbf{64 Steps} \\
\midrule
Walker2D   & 4083(68.8\%) & \textbf{4411(84.6\%)}$\uparrow$ & 4239(77.6\%)$\downarrow$ & 4125(71.2\%)$\downarrow$ & 4042(66.6\%)$\downarrow$ \\
\addlinespace
Adroit Pen Sparse & 2.47(86.2\%) & \textbf{2.94(87.6\%)}$\uparrow$ & 2.35(86.6\%)$\downarrow$ & 1.66(84.8\%)$\downarrow$ & 1.56(85\%)$\downarrow$ \\
\addlinespace
Humanoid Standup & 337{,}122(76.4\%) & 370{,}789(89.6\%)$\uparrow$ & \textbf{373{,}442(91.2\%)}$\uparrow$ & 365{,}999(88.6\%)$\downarrow$ & 365{,}868(88.4\%)$\downarrow$ \\
\bottomrule
\end{tabularx}

\caption{Average reward (success rate) comparison across inference steps on different tasks. The boldface entries denote the best performance for each task. 
$\uparrow$ ($\downarrow$) indicates that the reward increases (decreases) compared with the previous step setting. }
\end{table*}


\section{Methodology}

This section is organised as follows. We first review the preliminaries of flow matching and its application to robotic policy learning. Next, we analyse the instability of multi-step inference, showing how the loss of Lipschitz continuity near $t=1$ leads to degraded performance. We then demonstrate that the learned velocity field tends to drift toward training actions at late times, resulting in localised overfitting. Finally, we introduce our Dense-Jump Flow Matching Policy, which combines a non-uniform (e.g., U-shaped) time scheduling scheme during training with a Dense-Jump integration strategy at inference to address these challenges.

\subsection{Preliminaries}
\subsubsection{Flow Matching}
Flow matching is a generative modeling framework that learns a time-dependent
velocity field $v(x,t)$ to transport samples from a simple source distribution $p_0$
to a complex target distribution $p_1$. Specifically, let $t \in [0,1]$ denote an
artificial time variable interpolating between the two distributions. The evolution of
a sample $x_t$ is governed by the ODE:
\begin{equation}
    \frac{dx}{dt} = v(x,t),
\end{equation}
such that $x_0 \sim p_0$ and $x_1 \sim p_1$. In practice, $v_\theta(x,t)$ is trained
to match the true velocity field $v^*(x,t)$ induced by a chosen path between $p_0$
and $p_1$. Once learned, $v_\theta$ defines a generative policy that can be integrated
forward in time to produce new samples.

\subsubsection{Flow Matching Policy for Robotics}
Let $\mathbf{o}\in\mathbb{R}^{d_o}$ denote observations and $\mathbf{a}\in\mathbb{R}^{d_a}$ be actions.
In robotics, we are interested in a conditional target distribution $p_1(\cdot\mid\mathbf{o})$,
representing expert actions given observations. We adopt a Gaussian prior
$p_0 \sim \mathcal{N}(0,I_{d_a})$ as the source, and define intermediate states
$\mathbf{a}_t$ along the path from $p_0$ to $p_1$.
Among many possible path choices, we use linear interpolation \cite{liu2022flow},
which is widely employed in robotics due to its simplicity, stable few-step inference,
and strong empirical performance:
\begin{equation}
    \mathbf{a}_t = (1-t)\,\mathbf{a}_0 + t\,\mathbf{a}_1, \qquad t \sim U(0,1).
\end{equation}
Under this coupling the FM target velocity is constant:
\begin{equation}
    v^*(\mathbf{a}_t,t,\mathbf{o}) = \tfrac{d}{dt}\mathbf{a}_t = \mathbf{a}_1 - \mathbf{a}_0.
\end{equation}

We then train a neural network to approximate this velocity field via regression:
\begin{equation}
    \min_{\theta}\;
    \mathbb{E}_{p(\mathbf{a}_t|\mathbf{o}),\; t\sim p(t)}
    \Big[\;\big\|v_\theta(\mathbf{a}_t,t,\mathbf{o}) - (\mathbf{a}_1-\mathbf{a}_0)\big\|^2\;\Big].
\end{equation}

At inference, given observation $\mathbf{o}$, we draw $\mathbf{a}_0 \sim p_0$ and
numerically integrate the learned ODE $v_\theta(\mathbf{a}_t,t,\mathbf{o})$ from
$t=0$ to $t=1$ to obtain the final action $\mathbf{a}_1$.

\subsection{Instability of Multi-Step Inference}
\subsubsection{Lipschitz Continuity and the Picard–Lindelöf Theorem}
A function $f:\mathbb{R}^d\to\mathbb{R}^m$ is \emph{Lipschitz continuous} on a
domain $D$ if there exists $L>0$ such that
\begin{equation}
    \|f(x)-f(y)\| \leq L\|x-y\|, \qquad \forall x,y\in D.
\end{equation}

The smallest such $L$ is the \emph{Lipschitz constant}. Lipschitz continuity is a
key condition in the Picard–Lindelöf theorem \cite{teschl2012ode}, which ensures
existence and uniqueness of ODE solutions:

\begin{theorem}[Picard–Lindelöf]
Let $v:U\subseteq \mathbb{R}^n\times\mathbb{R}\to\mathbb{R}^n$
be continuous in $t$ and locally Lipschitz in $x$.
Then, for any $(t_0,x_0)\in U$, the initial value problem
\begin{equation}
    \frac{dx}{dt}=v(x,t), \quad x(t_0)=x_0,
\end{equation}
admits a unique solution in a neighborhood of $t_0$.
\end{theorem}

Thus, continuity in $t$ ensures smooth temporal evolution, while local Lipschitz
continuity in $x$ prevents trajectories starting from nearby states from diverging
arbitrarily. When these assumptions are violated, uniqueness and stability of the
ODE solution can no longer be guaranteed.

\subsubsection{Instability in Flow Matching Inference}
In the linear FM setting with $p_0 \sim \mathcal{N}(0,I_{d_a})$, the true velocity field is
\begin{equation}
    v^*(x,t) = \mathbb{E}\!\left[\tfrac{d}{dt}x_t \;\middle|\; x_t=x\right]
    = -\frac{x}{1-t}.
\end{equation}
For any $x,y\in\mathbb{R}^d$,
\begin{equation}
    \|v^*(x,t)-v^*(y,t)\| = \frac{\|x-y\|}{1-t}.
\end{equation}

so the minimal Lipschitz constant is
\begin{equation}
    L(t) = \frac{1}{1-t}.
\end{equation}

As $t \to 1^{-}$, we have $L(t) \to \infty$. Consequently, the local Lipschitz
assumption in the Picard–Lindelöf theorem fails near $t=1$, and the ODE
solutions lose their uniqueness and stability guarantees. Intuitively, the closer
the trajectory evolves to $t=1$, the more sensitive it becomes to perturbations
in initial conditions and numerical errors, causing instability in integration.
This explains the counterintuitive empirical finding that increasing the number
of inference steps often worsens policy performance: additional steps expose the trajectory more frequently to regions where the dynamics are nearly
non-Lipschitz, amplifying errors and degrading generalisation.

\begin{figure*}[t]
    \centering
    \begin{subfigure}[b]{0.32\linewidth}
        \centering
        \includegraphics[width=\linewidth]{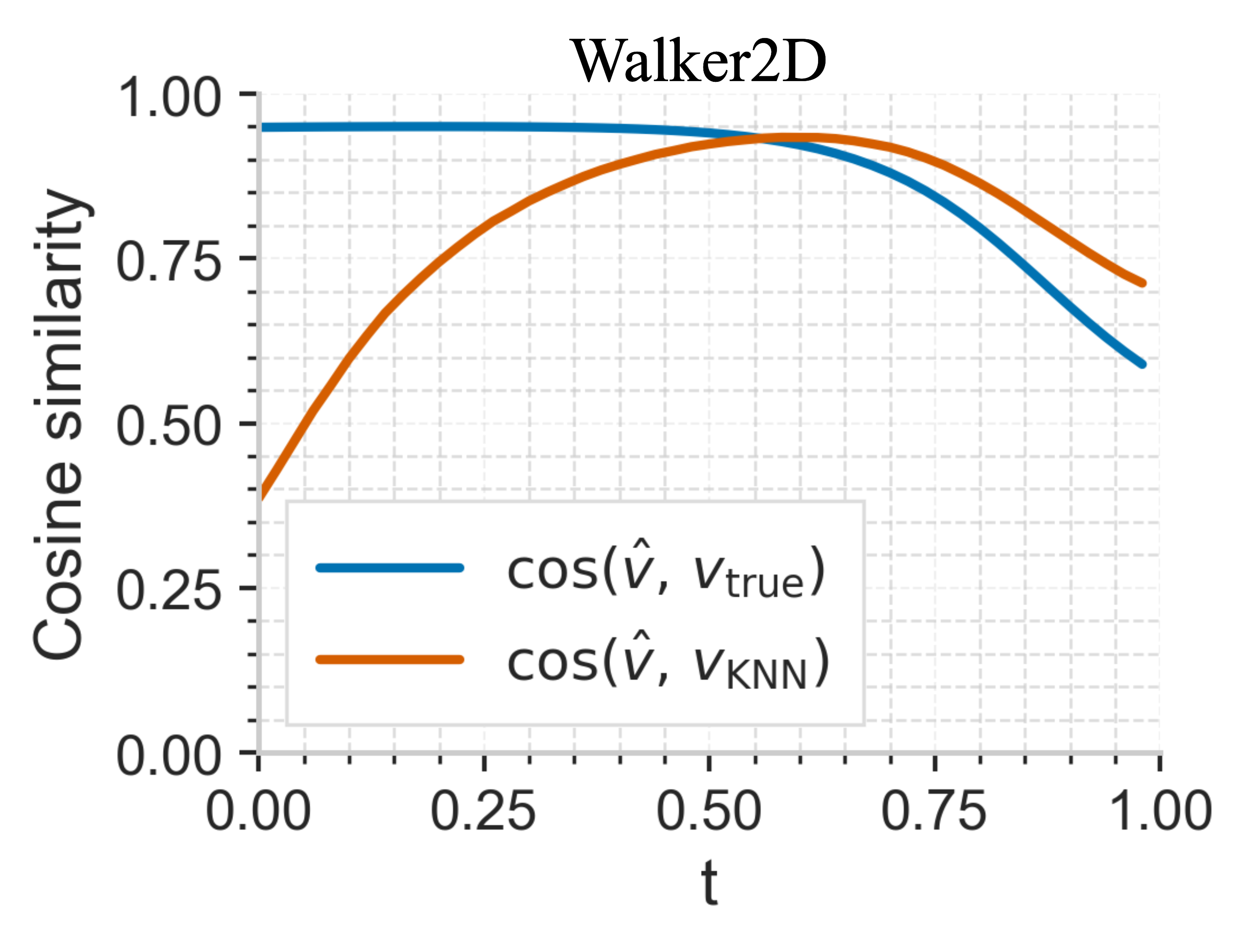}
        \label{fig:env:walker2d}
    \end{subfigure}
    \hfill
    \begin{subfigure}[b]{0.32\linewidth}
        \centering
        \includegraphics[width=\linewidth]{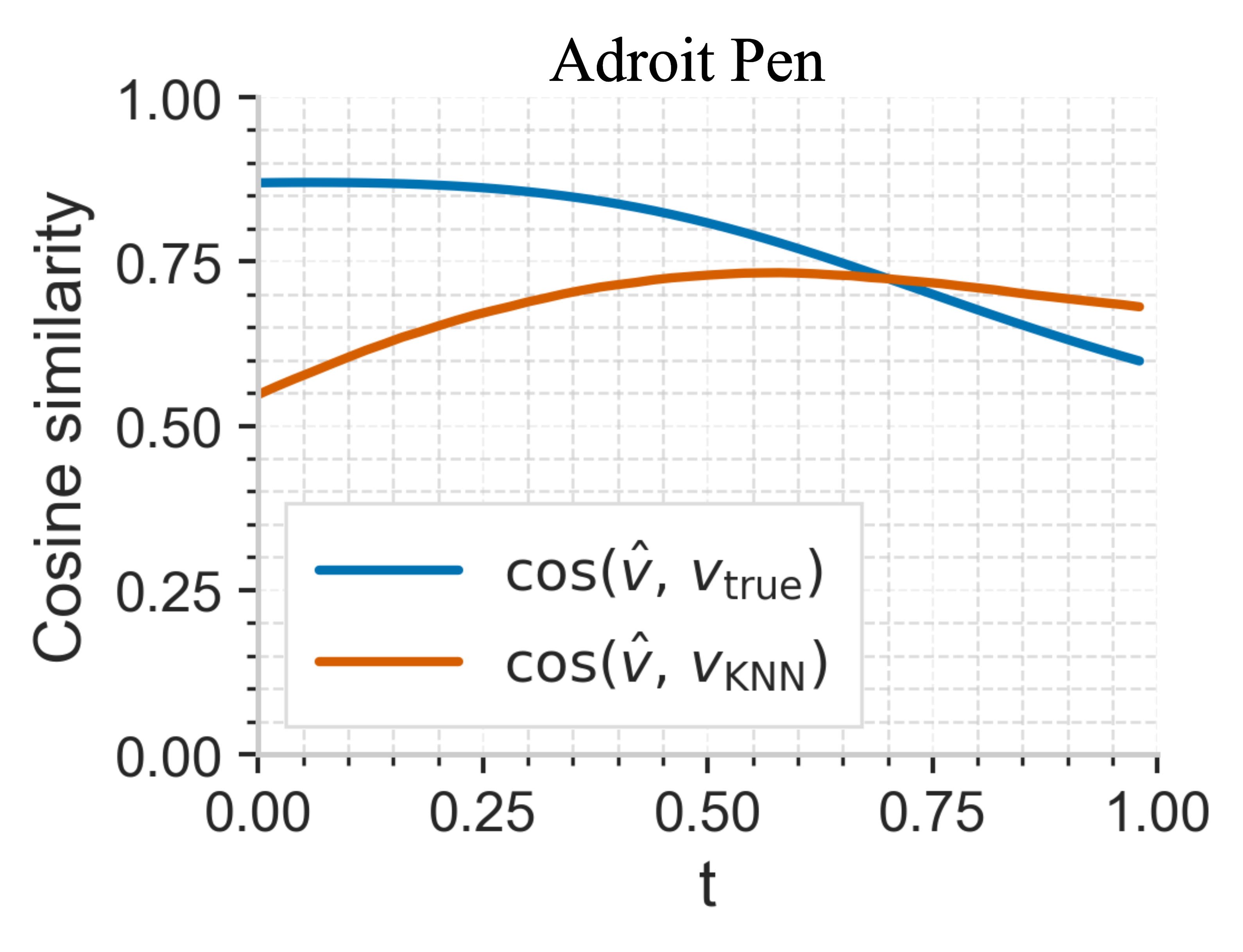}
        \label{fig:env:humanoid}
    \end{subfigure}
    \hfill
    \begin{subfigure}[b]{0.32\linewidth}
        \centering
        \includegraphics[width=\linewidth]{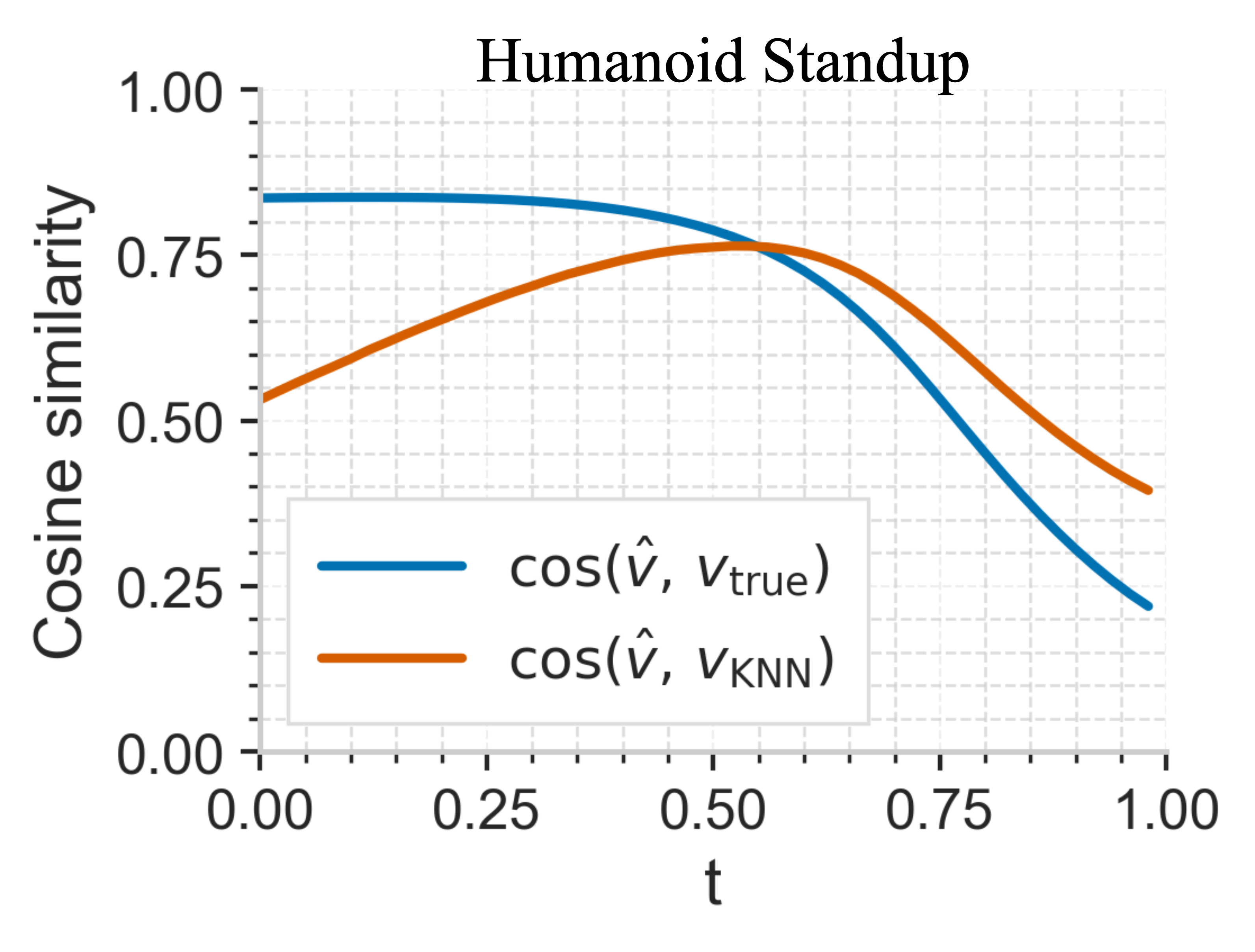}
        \label{fig:env:adroit}
    \end{subfigure}
    \caption{
    Cosine similarity between the learned velocity $\hat{\text{v}} = v_\theta(\mathbf{a}_t, t, \mathbf{o})$ and (1) the ground-truth expert velocity $\text{v}_{\text{true}} = \mathbf{a}_1|\mathbf{o} - \mathbf{a}_0$, and (2) the nearest training action velocity $\text{v}_{\text{KNN}} = \mathbf{a}_{\text{KNN}} - \mathbf{a}_0$, as a function of integration time $t$. Results are shown for Walker2D, Adroit Pen Sparse, and Humanoid Standup from left to right. Both similarities degrade as $t \to 1$, indicating universal performance decline at late times. Notably, in the mid-to-late stages, alignment with the nearest training action $\cos(\hat{\text{v}}, \text{v}_{\text{KNN}})$ exceeds that with the ground truth $\cos(\hat{\text{v}}, \text{v}_{\text{true}})$, demonstrating that the learned velocity field overfits to training actions rather than generalising to the true expert action. This highlights a localised overfitting phenomenon in flow matching policies at mid-to-late integration times.
    }
    \label{fig:overfitting}
\end{figure*}

\vspace{3pt}


\subsection{Late Time Flow Trajectories Drifting Toward Training Actions}

Recent research \cite{bertrand2025closed} demonstrates that the generalisation capacity of FM is predominantly established at early times. This limitation is less problematic in image generation, as the consequence is merely that generated images may closely resemble those in the training set. The situation, however, is markedly different in robotics. This is because both robotic state and action spaces are continuous, making exact repetitions of states virtually impossible in either simulation or the real world. Consequently, a policy that reproduces training-like actions can result in substantial performance degradation.

We further substantiate this claim in robotic tasks. For Walker2D, Adroit Pen, and Humanoid Standup, we independently construct a $K$-nearest neighbours (KNN) index over the expert action set, allowing us to retrieve an input’s nearest training action in action space. For any given state $\mathbf{o}$ in the remaining expert dataset and a generated action $\mathbf{a}_t$ at time step $t$, we use KNN to retrieve its nearest neighbour $\mathbf{a}_{\text{KNN}}$. We treat the expert action $\mathbf{a}_1|\mathbf{o}$ as ground truth to compute $\text{v}_{\text{true}} = \mathbf{a}_1|\mathbf{o} - \mathbf{a}_0$. In addition, we compute $\text{v}_{\text{KNN}} = \mathbf{a}_{\text{KNN}} - \mathbf{a}_0$ and $\hat{\text{v}} = v_\theta(\mathbf{a}_t, t, \mathbf{o})$, which represent the velocities induced by the KNN retrieved action and the FM generated action, respectively. The resulting cosine similarities between them are shown in Fig.~\ref{fig:overfitting}. We highlight two key observations across three robotics tasks, i.e., Walker2D, Adroit Pen, and Humanoid Standup:
\begin{itemize}
    \item Both $\cos(\hat{\text{v}}, \text{v}_{\text{KNN}})$ and $\cos(\hat{\text{v}}, \text{v}_{\text{true}})$ decrease as $t \to 1$, showing a universal degradation in the performance of FM policies.
    \item In the mid-to-late stages, $\cos(\hat{\text{v}}, \text{v}_{\text{KNN}})$ exceeds $\cos(\hat{\text{v}}, \text{v}_{\text{true}})$, indicating that the learned velocity field aligns closer to certain training actions than to the ground truth. 
\end{itemize}

This indicates that the learned velocity field exhibits overfitting localised to specific regions of the time interval, often reflecting the intrinsic difficulty of the task. This phenomenon differs fundamentally from conventional overfitting in standard machine learning frameworks, as it arises only in partial segments of the temporal domain. Moreover, we find that traditional regularisation techniques cannot effectively mitigate the challenge, underscoring the need for FM-specific remedies.

\subsection{Dense-Jump Flow Matching Policy}
We adopt a non-uniform (e.g., U-shaped) time scheduling scheme that down-weights interior times, where localised overfitting is most pronounced, and allocates extra probability mass near both endpoints. This approach strengthens early conditioning signals and stabilises the terminal jump used at inference.

\subsubsection{Non-Uniform Time Scheduling}

\textbf{Early-time anchoring ($t\!\approx\!0$).}
    When $t$ is very small, $\mathbf{a}_t \approx \mathbf{a}_0$, action samples remain dominated by Gaussian source noise. The target displacement $(\mathbf{a}_1 - \mathbf{a}_0)$ is therefore weakly expressed in $\mathbf{a}_t$, yielding a poor signal-to-noise ratio (SNR) for direct velocity regression. In this regime the gradient signal flowing through $v_\theta$ is predominantly condition-led: the observation embedding (or state encoding) supplies the informative structure while the action component is largely stochastic. By deliberately allocating extra sampling mass near $t \approx 0$, we: (i) encourage the network to align its early-time feature space tightly with $\mathbf{o}$, improving conditional faithfulness; (ii) reduce reliance on mid-interval timesteps where memorisation of specific training actions emerges (refer to Fig. \ref{fig:overfitting}); (iii) obtain lower-variance gradient estimates because the model learns a coarse, observation-conditioned directional prior before refining magnitudes later; and (iv) implicitly regularise the Jacobian $\nabla_{\mathbf{a}} v_\theta$ in a regime where the underlying flow is smooth, yielding better stability for few-step or one-step inference. This early anchoring lessens the drift toward nearest-neighbour training actions observed at intermediate $t$, and synergises with the late-time allocation below to stabilise the Dense-Jump update.

\begin{figure}[htbp]
    \centering
    \includegraphics[width=1.0\linewidth]{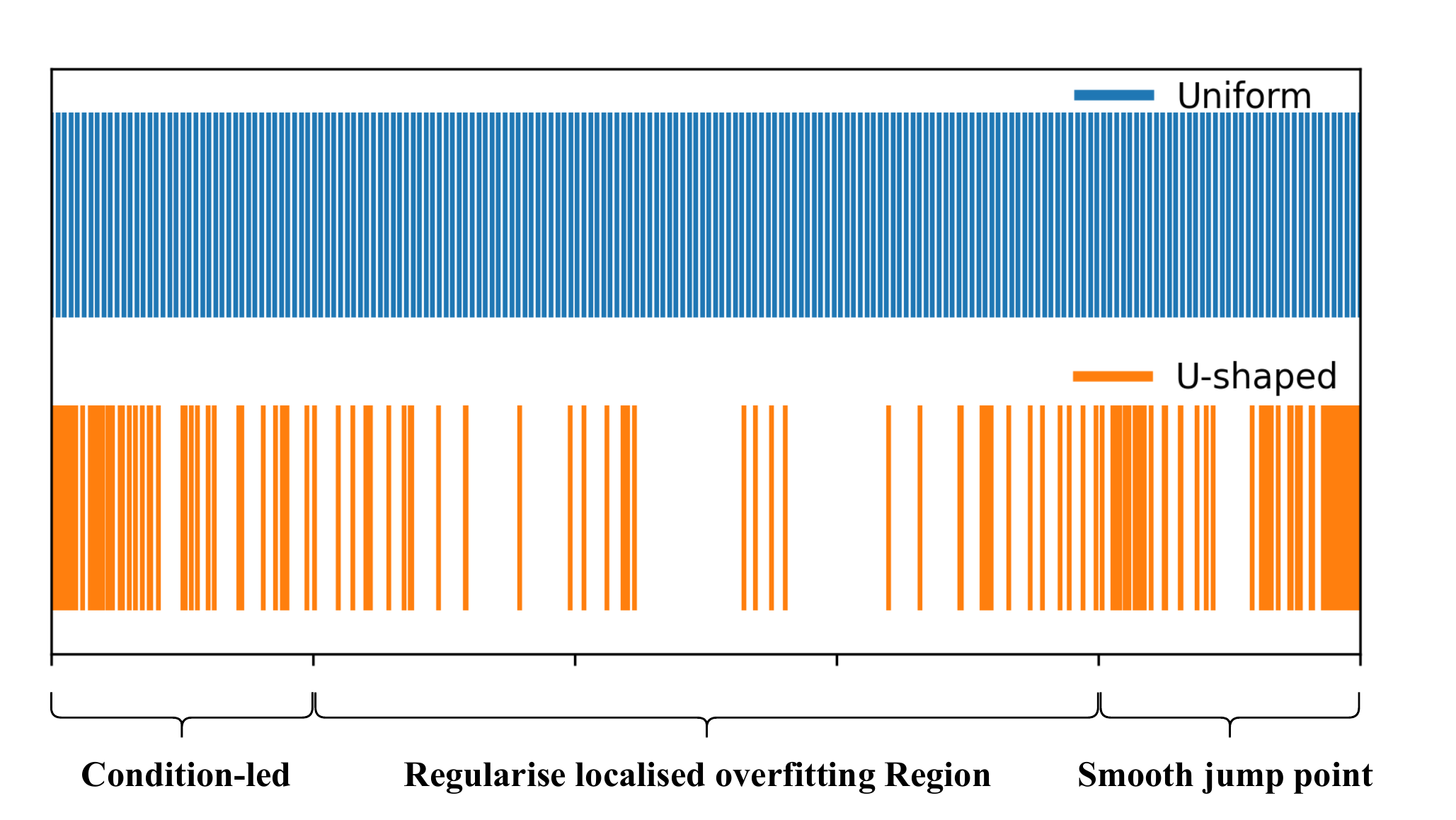}
    \caption{Comparison of time sampling strategies for flow matching training. The uniform sampling (top) allocates equal probability mass across all time steps $t \in [0,1]$, while the U-shaped sampling (bottom) concentrates additional probability mass at early times ($t \approx 0$) and late times ($t \approx 1$) with reduced sampling in the intermediate region. The U-shaped distribution enables more intensive training coverage in the condition-led region (near $t=0$) and the critical terminal region (near $t=1$) where accurate velocity field learning is essential for stable Dense-Jump inference, while maintaining regularisation against localised overfitting in the intermediate time steps.}
    \label{fig:beta}
\end{figure}

\textbf{Beta scheduling}: One simple yet effective way to achieve U-shaped non-uniform sampling is through the well-defined Beta distribution $t\!\sim\!\mathrm{Beta}(\alpha,\alpha)$, $\alpha \in (0,1)$ with density
\begin{equation}
    w_\alpha(t)=\tfrac{t^{\alpha-1}(1-t)^{\alpha-1}}{B(\alpha,\alpha)},
\end{equation}
\noindent where $B(\alpha,\alpha)$ is the Beta function. The parameter $\alpha$ controls the degree of non-uniformity: smaller $\alpha$ produces a more pronounced U-shape, concentrating probability mass near both endpoints $t=0$ and $t=1$ (Fig.~\ref{fig:beta}).

\subsubsection{Late-Time Smoothing for Dense-Jump Inference}
Our Dense-Jump strategy executes a single terminal jump from $t_{\text{jump}}$ to $1$. 
Let $\Delta = 1 - t_{\text{jump}}$ and consider the non-autonomous dynamics
$\dot{\mathbf{a}}_t = v_\theta(\mathbf{a}_t, t, \mathbf{o})$.
A one-step Taylor expansion about $t_{\text{jump}}$ gives
\begin{equation}
    \mathbf{a}_1
        = \mathbf{a}_{t_{\text{jump}}}
            + \Delta\, v_\theta(\mathbf{a}_{t_{\text{jump}}}, t_{\text{jump}}, \mathbf{o})
            + \tfrac{1}{2}\Delta^2 \kappa_\theta(\xi)
            + O(\Delta^3),
\end{equation}
\noindent
where
\begin{equation}
    \kappa_\theta(t)
        = \partial_t v_\theta(\mathbf{a}_t,t,\mathbf{o})
          + \big[\nabla_{\!\mathbf{a}} v_\theta(\mathbf{a}_t,t,\mathbf{o})\big]\,
            v_\theta(\mathbf{a}_t,t,\mathbf{o}),
\end{equation}
where $\xi \in [t_{\text{jump}},1]$ and $\nabla_{\!\mathbf{a}}$ denotes the Jacobian with respect to $\mathbf{a}$. Thus the local truncation error of the terminal jump scales as
$\tfrac{1}{2}\Delta^2 \|\kappa_\theta(\xi)\| + O(\Delta^3)$.

Without adequate training coverage near $t \approx 1$, the neural network used to learn the policy tends to extrapolate into a high-gradient, high-curvature regime, often inducing oscillatory or unstable behaviour. By allocating additional probability mass to late times, we essentially expose the network more frequently to this regime, leading to more sustained supervision of the neural networks. This will eventually lead the network to favour smoother (low frequency) over highly oscillatory (high frequency) fits, acting as an implicit data-driven regulariser~\cite{guo2025ugod}. The resulting velocity field in the terminal region is therefore less oscillatory, effectively reducing the curvature term $\|\kappa_\theta\|$ and the jump error in Dense-Jump inference, even though the intrinsic Lipschitz blow-up as $t \to 1$ remains unavoidable.





\subsubsection{Dense-Jump Integration}
To address performance degradation at inference, we propose the Dense-Jump ODE solver. We select a switch time $t_{\text{jump}}\!\in\!(0,1)$ and allocate the step budget $N$ to (i) $(N{-}1)$ uniform explicit Euler steps over the stable interval $[0,t_{\text{jump}}]$ with step size $\Delta t = t_{\text{jump}}/(N{-}1)$, followed by (ii) a single terminal jump from $t_{\text{jump}}$ to $1$. This concentrates numerical effort where the velocity field remains moderately Lipschitz, and avoids repeatedly traversing the late-time region where the effective Lipschitz constant $L(t)\!\sim\!(1-t)^{-1}$ (and hence local amplification of discretisation and estimation errors) explodes. The cumulative local truncation error of the dense phase scales as $O(t_{\text{jump}}\Delta t)=O\!\big(t_{\text{jump}}^{2}/(N{-}1)\big)$ under standard Euler assumptions. For the terminal jump we write a one–step Taylor expansion:
\begin{equation}
\begin{aligned}
\mathbf{a}_1
&= \mathbf{a}_{t_{\text{jump}}}
    + (1-t_{\text{jump}})\, v_{\hat{\theta}}(\mathbf{a}_{t_{\text{jump}}}, t_{\text{jump}}, \mathbf{o}) \\
&\quad + \tfrac{1}{2}(1-t_{\text{jump}})^2 \kappa_{\hat{\theta}}(\xi)
    + O\!\big((1-t_{\text{jump}})^3\big).
\end{aligned}
\end{equation}
where $\kappa_{\hat{\theta}}(t)=\partial_t v_{\hat{\theta}} + \big(\nabla_{\mathbf{a}} v_{\hat{\theta}}\big) v_{\hat{\theta}}$ and $\xi\in[t_{\text{jump}},1]$, and $\hat{\theta}$ indicates the model training is complete. Thus the jump contributes $O\!\big((1-t_{\text{jump}})^2 \|\kappa_{\hat{\theta}}\|\big)$ error. Training with the U-shaped $t\sim\mathrm{Beta}(\alpha,\alpha)$ ($\alpha<1$) supplies dense supervision near both $t{=}0$ (improving conditional anchoring) and $t{=}1$ (reducing curvature $\|\kappa_{\hat{\theta}}\|$ via repeated exposure), empirically lowering the terminal error constant. Compared to a uniform $N$-step Euler schedule over $[0,1]$, which accrues many small steps precisely even when $L(t)$ is largest (near non-Lipschitz), Dense-Jump replaces the $O\!\big((1-t_{\text{jump}})^2 \|\kappa_{\hat{\theta}}\|\big)$ high-sensitivity updates by a single controlled extrapolation. Because robotic control typically tolerates small terminal perturbations (task reward is often insensitive to infinitesimal action deviations), the stability gained by skipping the near–non-Lipschitz tail outweighs the mild increase in one-step truncation error, yielding a better robustness–accuracy trade-off under fixed compute. Details of the overall training and inference procedure are summarised in Algorithm~\ref{alg:djfm_main}.

\begin{algorithm}[t]
\caption{Dense-Jump Flow Matching Policy}
\label{alg:djfm_main}
\small
\begin{algorithmic}[1]
\STATE \textbf{Inputs:} Condition $\mathbf{o}$; velocity model $v_\theta(\cdot,t,\mathbf{o})$; jump time $t_{\text{jump}} \in (0,1)$; number of flow steps $N$.
\STATE \textbf{Coupling:} Draw $(\mathbf{a}_1,\mathbf{o}) \sim p_{\text{data}}$, and $\mathbf{a}_0 \sim p_0$.
\vspace{4pt}
\STATE \textbf{Training:}
\[
\hat{\theta} \;=\; \arg\min_\theta\;
\mathbb{E}\Big[\big\| v_\theta(\mathbf{a}_t,t,\mathbf{o}) - (\mathbf{a}_1-\mathbf{a}_0) \big\|_2^2 \Big],
\]
with $t \sim \mathrm{Beta}(\alpha,\alpha)$, $\;\alpha \in (0,1)$.
\vspace{4pt}
\STATE \textbf{Deployment:}
\begin{enumerate}[label*=\arabic*., leftmargin=2em]
    \item Set $\Delta t \leftarrow t_{\text{jump}} / {(N-1)}$.
    \item Initialize $\mathbf{a} \leftarrow \mathbf{a}_0$, $\; t \leftarrow 0$.
    \item \textbf{For} $i=1$ to $N-1$:
    \begin{enumerate}[label*=\arabic*., leftmargin=2em]
        \item $\mathbf{a} \leftarrow \mathbf{a} + \Delta t\, v_{\hat{\theta}}(\mathbf{a}, t, \mathbf{o})$.
        \item $t \leftarrow t + \Delta t$.
    \end{enumerate}
    \item \textbf{Terminal jump:}
    \[
    \mathbf{a}_1 \;=\; \mathbf{a} + (1 - t_{\text{jump}})\, v_{\hat{\theta}}(\mathbf{a}, t_{\text{jump}}, \mathbf{o}).
    \]
\end{enumerate}
\end{algorithmic}
\end{algorithm}

\begin{figure*}[h]
    \centering
    \begin{subfigure}[b]{0.32\linewidth}
       \centering
        \includegraphics[width=\linewidth]{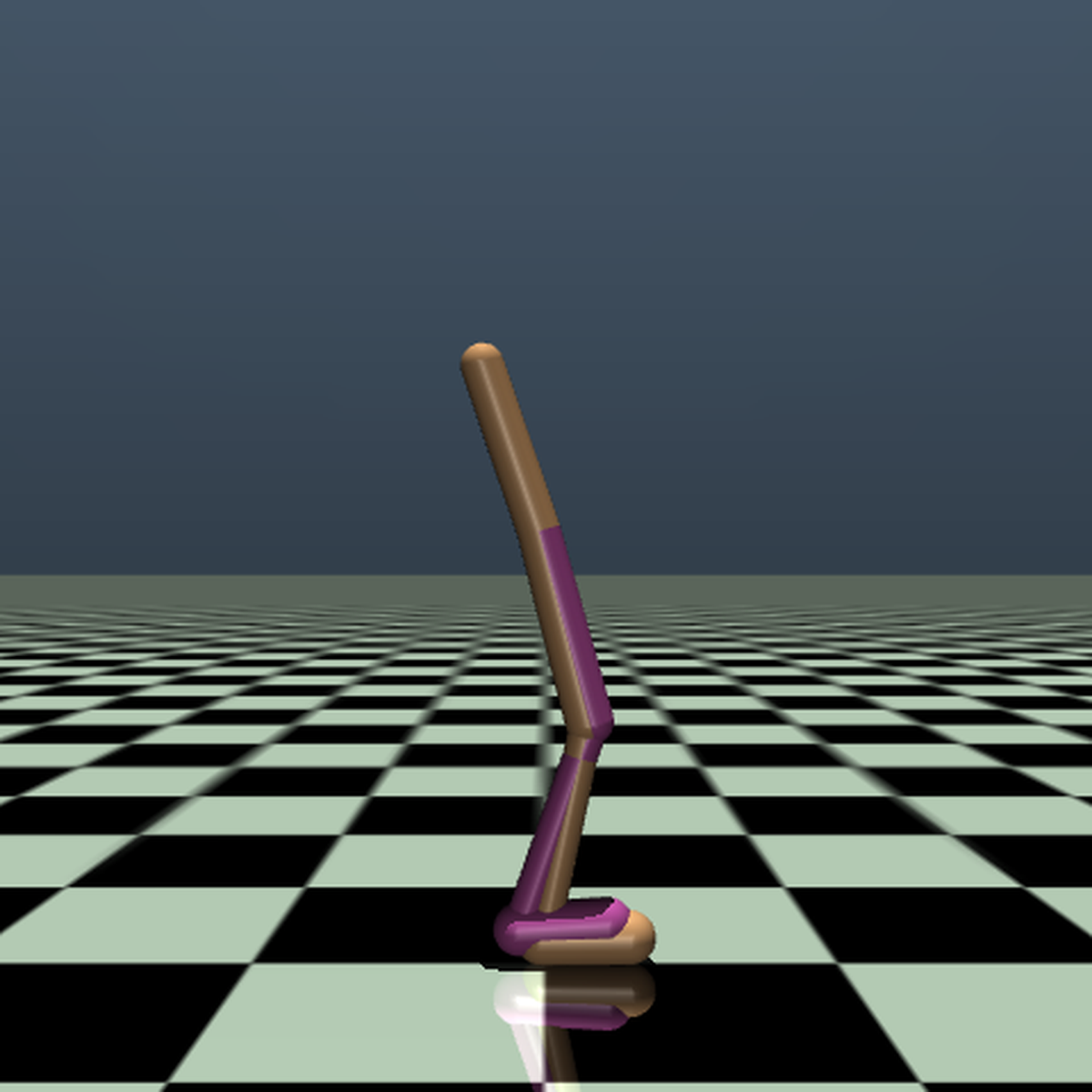}
        \caption{Walker2D}
        \label{fig:env:walker2d}
    \end{subfigure}
    \hfill
    \begin{subfigure}[b]{0.32\linewidth}
       \centering
        \includegraphics[width=\linewidth]{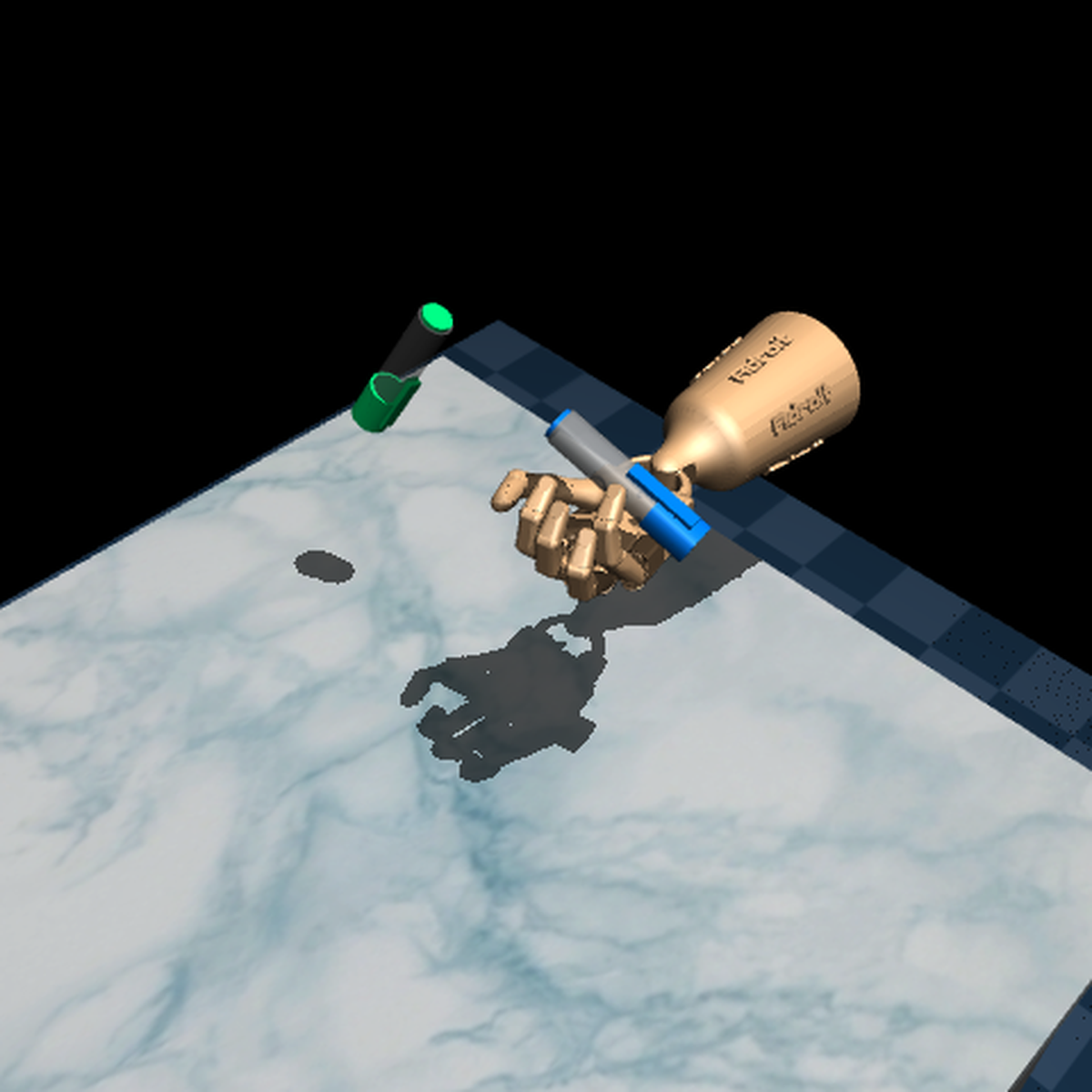}
       \caption{Adroit Pen}
        \label{fig:env:humanoid}
    \end{subfigure}
   \hfill
    \begin{subfigure}[b]{0.32\linewidth}
       \centering
        \includegraphics[width=\linewidth]{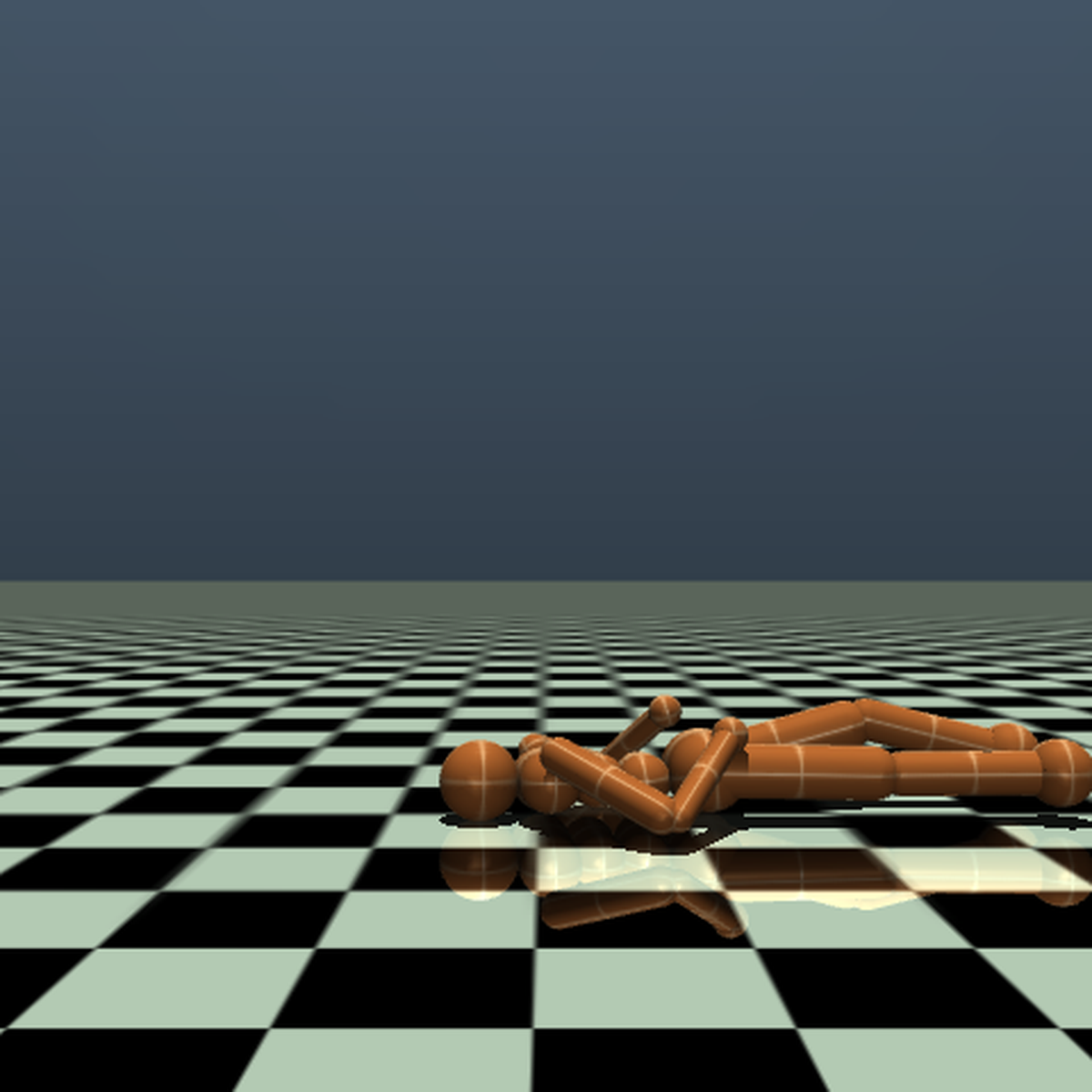}
        \caption{Humanoid Standup}
        \label{fig:env:adroit}
    \end{subfigure}
    \caption{Three simulation benchmarks used in our experiments.}
   \label{fig:envs}
\end{figure*}

\section{Experiments}
\subsection{Simulation Environment and Expert Data}
We evaluate the proposed Dense-Jump flow matching on three widely accepted simulation benchmarks spanning a range of complexity levels. These environments can be accessed through the Gymnasium API~\cite{towers2023gymnasium}, and our implementation is open-sourced at \url{https://github.com/DenseJumpFM/DenseJump_FlowMatching}. 

\textbf{Walker2D-v5} (MuJoCo). The agent receives a 17-dimensional observation vector (8 joint positions and 9 joint velocities) and outputs 6-dimensional torque actions applied at the joints. The objective is to maintain stable forward locomotion without falling.  

\textbf{HumanoidStandup-v5} \cite{tassa2012synthesis} (MuJoCo). The agent observes a 348-dimensional state vector (including joint positions, velocities, and contact-related information) and produces 17-dimensional torque actions. The objective is to train the humanoid to stand up from a prone configuration and sustain an upright stance.  

\textbf{AdroitHandPenSparse-v1} \cite{Rajeswaran-RSS-18} (Robotics Suite). The 24-DoF Shadow Hand issues 24-dimensional joint angle commands conditioned on a 45-dimensional observation vector comprising hand state, pen pose, and target orientation. The goal is to re-orient a pen so that its pose matches a randomly initialised target orientation within a specified tolerance. According to the official documentation, the episode should terminate upon success; however, this behaviour is inconsistent with the released implementation. We therefore introduce a lightweight custom wrapper to enforce the documented termination semantics. 



\begin{table*}[!ht]
\centering
\caption{Comparison of FM-DJ$\beta$ with flow matching and diffusion policies across 3 tasks under different inference steps. `—' denotes failed configurations under the given step budget. Each entry is in the form of `reward(success rate)'. Best results are \textbf{highlighted}.}
\label{tab:inference_main}
\begin{tabularx}{\textwidth}{llccccc}
\hline
\textbf{Task} & \textbf{Method} & \textbf{1 step} & \textbf{2 steps} & \textbf{4 steps} & \textbf{16 steps} & \textbf{64 steps} \\
\hline
\multirow{3}{*}{Walker2D} 
 & Diffusion & - & - & - & 1,732(9\%) & 2,738(25\%) \\
 & Vanilla FM & 4,083(68.8\%) & 4,411(84.6\%) & 4,239(77.6\%) & 4,125(71.2\%) & 4,042(66.6\%) \\
 & FM-DJ$\beta$ (Ours) & $\mathbf{4,410(82.2\%)}$ & $\mathbf{4,543(88.2\%)}$ & $\mathbf{4,542(86.4\%)}$ & $\mathbf{4,634(89.8\%)}$ & $\mathbf{4,609(86.8\%)}$ \\
\hline
\multirow{3}{*}{Adroit Pen Sparse} 
 & Diffusion   & -10.62(34\%) & -12.41(27\%)&-9.64(38\%) & -7.80(44\%) & -6.53(49\%)\\
 & Vanilla FM & 2.47(86.2\%) & 2.66(87.2\%) & 2.35(86.6\%) & 1.66(84.8\%) & 1.56(85\%) \\
 & FM-DJ$\beta$ (Ours) &  $\mathbf{2.88(87.2\%)}$ & $\mathbf{3.15(88.4\%)}$ & $\mathbf{3.29(89.2\%)}$ & $\mathbf{3.00(88.2\%)}$ & $\mathbf{2.74(86.4\%)}$ \\
\hline
\multirow{3}{*}{Humanoid Standup} 
 & Diffusion & -& - &- & - & -\\
 & Vanilla FM & 337{,}122(76.4\%) & $\mathbf{370{,}789(89.6\%)}$ & 373{,}442(91.2\%) & 365{,}999(88.6\%) & 365{,}868(88.4\%) \\
 & FM-DJ$\beta$ (Ours) & $\mathbf{346{,}363(80\%)}$ & 370{,}099(89\%) & $\mathbf{375{,}993(91.4\%)}$ & $\mathbf{374{,}369(91\%)}$ & $\mathbf{372{,}547(90.4\%)}$ \\
\hline
\end{tabularx}
\end{table*}

\subsection{Experiment Setup}

All experiments are conducted on a single NVIDIA A100 GPU. Unless otherwise stated, results are reported as the mean over 5 random seeds during training and averaged over 100 simulation rollouts at inference. The training datasets consist of the first 8 episodes (8{,}000 steps) for Walker2D; 30 episodes (6{,}000 steps) for Adroit Pen; and 100 episodes (100{,}000 steps) for Humanoid Standup. We train all policy networks for 5{,}000 epochs with a batch size of 128 on Walker2D and Adroit Pen, and for 2{,}000 epochs with a batch size of 1{,}024 on Humanoid Standup, saving checkpoints every 100 epochs. For reporting, we use the checkpoint that achieves the best validation performance.
For time scheduling, we adopt a simple U-shaped, non-uniform sampling \(t \sim \mathrm{Beta}(0.2,\,0.2)\) during training and use a Dense-Jump solver at inference with a fixed jump point \(t_{\text{jump}} = 0.5\).

The three tasks are shown in Fig.~\ref{fig:envs}. We compare our method with two strong baselines: (i) the original diffusion policy with 1, 2, 4, 16, and 64 inference steps; and (ii) the vanilla flow matching policy trained with uniform time sampling and evaluated under the same inference step budgets.

\section{Results and Analysis}
Table~\ref{tab:inference_main} summarises performance across step budgets \{1, 2, 4, 16, 64\} on the three benchmarks. Overall, we can see that the proposed method FM-DJ$\boldsymbol{\beta}$ maintains robust performance across the entire step budget range while achieving superior one-step inference. In comparison, vanilla FM achieves strong performance at low step counts (2 - 4 steps) and exhibits systematic degradation as integration steps increase. Diffusion policies either fail outright at low step counts or require many steps to achieve modest results. Some key observations and insights are as follows:

\begin{table*}[bp]
\centering
\caption{Ablation results. Vanilla FM: uniform time sampling + uniform Euler. FM-$\beta$: U-shaped (Beta) time sampling only. FM-DJ: Dense-Jump inference only. FM-DJ$\beta$: both. FM-$\beta$ improves one/few-step accuracy but becomes brittle with many steps. FM-DJ stabilises multi-step inference but yields smaller single-step gains. Their combination (FM-DJ$\beta$) consistently attains the best or tied-best performance across tasks and step budgets, showing the two components are complementary. Best results in bold.}
\label{tab:ablation}
\begin{tabularx}{\textwidth}{llccccc}
\hline
\textbf{Task} & \textbf{Method} & \textbf{1 step} & \textbf{2 steps} & \textbf{4 steps} & \textbf{16 steps} & \textbf{64 steps} \\
\hline
\multirow{4}{*}{Walker2D}
 & Vanilla FM & 4,083(68.8\%) & 4,411(84.6\%) & 4,239(77.6\%) & 4,125(71.2\%) & 4,042(66.6\%) \\
 & FM-$\beta$ & \textbf{4,410(82.2\%)} & \textbf{4,543(88.2\%)} & 4,344(76.8\%) & 3,402(44.6\%) & 3,649(44.2\%) \\
 & FM-DJ & 4,083(68.8\%) & 4,411(84.6\%) & 4,276(82.2\%) & 4,438(87.4\%) & 4,378(85\%) \\
 & FM-DJ$\beta$ (Ours) & \textbf{4,410(82.2\%)} & \textbf{4,543(88.2\%)} & \textbf{4,542(86.4\%)} & \textbf{4,634(89.8\%)} & \textbf{4,609(86.8\%)} \\
\hline
\multirow{4}{*}{Adroit Pen Sparse}
 & Vanilla FM & 2.47(86.2\%) & 2.66(87.2\%) & 2.35(86.6\%) & 1.66(84.8\%) & 1.56(85\%) \\
 & FM-$\beta$ & \textbf{2.88(87.2\%)} & \textbf{3.15(88.4\%)} & 2.44(87.2\%) & 1.44(84.6\%) & 1.14(83\%) \\
 & FM-DJ & 2.47(86.2\%) & 2.66(87.2\%) & 2.53(86.4\%) & 2.64(87.2\%) & 2.28(86\%) \\
 & FM-DJ$\beta$ (Ours)& \textbf{2.88(87.2\%)} & \textbf{3.15(88.4\%)} & \textbf{3.29(89.2\%)} & \textbf{3.00(88.2\%)} & \textbf{2.74(86.4\%)} \\
\hline
\multirow{4}{*}{Humanoid Standup}
 & Vanilla FM & 337{,}122(76.4\%) & \textbf{370{,}789(89.6\%)} & 373{,}442(91.2\%) & 365{,}999(88.6\%) & 365{,}868(88.4\%) \\
 & FM-$\beta$ & \textbf{346{,}363(80\%)} & 370{,}099(89\%) & 372{,}212(90.8\%) & 357{,}604(85\%) & 348{,}782(82\%) \\
 & FM-DJ & 337{,}122(76.4\%) & \textbf{370{,}789(89.6\%)} & 368{,}441(89.4\%) & 370{,}721(89.8\%) & 368{,}847(88.6\%) \\
 & FM-DJ$\beta$ (Ours) & \textbf{346{,}363(80\%)} & 370{,}099(89\%) & \textbf{375{,}993(91.4\%)} & \textbf{374{,}369(91\%)} & \textbf{372{,}547(90.4\%)} \\
\hline
\end{tabularx}
\end{table*}

\textbf{FM-DJ$\beta$} yields significant improvements in both one-step inference performance and peak performance. The one-step results improve substantially across tasks: on Walker2D, the single-step average reward increases about 8\% (from 4{,}083 to 4{,}410), accompanied by a success rate rise from 68.8\% to 82.2\%. On Adroit Pen, the one-step reward climbs from 2.47 to 2.88 (approximately +16.6\%), with success rate improving from 86.2\% to 87.2\%. Similarly, on Humanoid Standup, the one-step reward improves from 337{,}122 to 345{,}363 (+2.4\%) and the success rate from 76.4\% to 80.0\%. These gains validate that FM-DJ$\beta$ can serve as an effective one-step policy without sacrificing success rates, directly addressing the one-step inference limitations of vanilla FM.

Critically, FM-DJ$\beta$ also advances the best performance achievable with multiple integration steps. On Walker2D, our method raises the peak average reward from 4{,}411 to 4{,}634, a 5.1\% increase over vanilla FM. For Adroit Pen, the maximum average reward is lifted from 2.66 to 3.29, which constitutes a substantial 23.7\% gain. While vanilla FM reaches its highest Humanoid Standup return of 373{,}442 at 4 steps, FM-DJ$\beta$ slightly surpasses this with 375{,}993 at the same step count. Notably, FM-DJ$\beta$ achieves these improvements without compromising single-step performance, reflecting how our Dense-Jump strategy avoids late-time instabilities and pushes the performance frontier forward.

\textbf{Vanilla FM} exhibits consistent step-count brittleness across all benchmarks. It follows a characteristic concave performance curve: results peak at 2--4 solver steps and then decline sharply as more steps are taken. For instance, in Walker2D, performance drops from 4{,}411 (84.6\% success) at 2 steps to 4{,}042 (66.6\% success) at 64 steps---an 8.4\% reward decrease accompanied by an 18 percentage-point drop in success rate. Adroit Pen (Sparse) shows even more severe degradation: the best reward of 2.66 at 2 steps plummets to 1.56 at 64 steps (approximately a 41\% reduction). Humanoid Standup also declines from 373{,}442 (91.2\% success) at 4 steps to 365{,}868 (88.4\%) at 64 steps. This systematic deterioration confirms that repeatedly integrating through late-time regions harms control quality, as accumulated errors in the learned velocity field compound over many small steps. These observations align with our theoretical analysis that the velocity field becomes non-Lipschitz as $t \to 1$, leading to unstable integrations and underscoring the importance of our proposed remedies.

\textbf{Diffusion} baselines reveal fundamental architectural limitations. The diffusion baseline consistently underperforms FM variants and exhibits severe step-count dependence. On Walker2D, diffusion completely fails for 1--4 steps, achieving only 1,732 (9\%) at 16 steps and 2,738 (25\%) at 64 steps. Adroit Pen shows strong inference step-count dependence: success rates remain low with 1--4 denoising steps and only improve at 16--64 steps. Most tellingly, on Humanoid Standup, diffusion fails to meet evaluation thresholds even with 64 steps.

These results establish that step robustness, not just peak performance, is essential for practical deployment where computational budgets vary. FM-DJ$\beta$ provides reliable performance across the entire step range, addressing a critical limitation of vanilla flow matching.

\subsection{Ablation Study}
We further investigate how the two components of FM-DJ$\beta$ (U-shaped time sampling and Dense-Jump integration) contribute to the overall performance. We consider two ablation variants: (i) FM-$\beta$, which employs U-shaped time sampling but uses standard uniform Euler integration at inference; and (ii) FM-DJ, which uses Dense-Jump integration but trains with uniform time sampling.

Table~\ref{tab:ablation} disentangles the contributions of the two components. 
{FM-$\beta$ (U-shaped time sampling) alone} boosts few-step accuracy but is brittle for large step budget case: e.g., Walker2D falls from 4,543 (88.2\%) at 2 steps to 3,402 (44.6\%) at 16 steps. 
FM-DJ (Dense-Jump) alone stabilises multi-step inference and helps to mitigate the performance degradation. In contrast,
By combining both non-uniform sampling and dense-jump, (FM-DJ$\beta$) delivers the best performance across different tasks: strong one-/few-step accuracy and robust multi-step performance, dominating all baselines across tasks and budgets. 

It is worth noting that, even though vanilla flow matching performs better with 2-step inference in the Humanoid Standup task, our method, FM-DJ$\beta$, still outperforms flow matching with dense jumps. This further demonstrates that our method achieves a synergistic effect, highlighting the coherence of our design choices.

\section{Conclusion and Future Work}
This work investigates why flow-matching robotic policies degrade when increasing inference steps. We attribute the decline to (i) the learned velocity field becoming effectively non-Lipschitz near $t \to 1$, amplifying numerical and modeling errors, and (ii) uniform time sampling can oversample in mid to late temporal intervals, causing the policy to memorise training actions that undermines its capacity to generalise. 
The proposed FM-DJ$\beta$ method addresses this by combining a non-uniform time schedule that strengthens supervision at the endpoints with a Dense-Jump integration scheme that concentrates computation in the stable early segment and replaces unstable multi-step inference with a single terminal jump. This produces higher returns, improved success rates, and markedly better robustness across step budgets.

Future directions include systematic research into Lipschitz-aware time sampling driven by online estimates of $\|\nabla_{\mathbf{a}} v_\theta\|$, which directly relates to the few/one step inference errors. This will facilitate a learning-based jump policy for robotics. We will also seek to deploy our policy in real-robotic scenarios with timing and actuation noises, to contribute to the robust rollouts of flow matching based policies in real-world settings.


\bibliographystyle{IEEEtran}
\bibliography{reference}
\end{document}